\documentclass[11pt]{article}
\usepackage[margin=1in]{geometry}
\usepackage{amsmath}
\usepackage{booktabs}
\usepackage{url}
\usepackage[colorlinks=true,linkcolor=blue,citecolor=blue,urlcolor=blue]{hyperref}
\usepackage{fontspec} 
\usepackage[UTF8,fontset=fandol]{ctex}
\usepackage{microtype}
\title{Grounding Language Models in Speech Does Not Fix ``AI Flavor'':\\
Calibrated LLM Judges, Goodhart Traps,\\
and a Preference-Tuning Fix}

\author{Dongwu Qian\\
\small Independent Researcher\\
\small \texttt{qiandongwu123@qq.com}\\
\small Code and data: RA-Eval v0.1, Zenodo DOI 10.5281/zenodo.24237533}
\date{November 2026}

\begin{document}
\maketitle

\begin{abstract}
Chinese outputs of large language models carry a perceptible ``AI flavor'': formulaic sentence frames (中文“不是……而是”“不仅……更”; lit.\ ``not\dots but rather'', ``not only\dots but even more''), stacked written-register connectives, repetition loops, and layout leakage. A popular hypothesis attributes this to text-only training, which lacks the inner-speech layer through which human writing passes; injecting speech or prosody signals into training should therefore reduce it. We test this hypothesis with controlled interventions on a 0.6B model, and all three fail: (i)~phoneme-interleaved continued pretraining yields no spokenness gain under text-matched controls while significantly increasing repetition ($+0.13$ $[+0.03,+0.23]$); (ii)~prosody-plan conditional generation reaches an adherence of only $0.257$ against a pre-registered $0.50$ gate; (iii)~a rule-based reward built on TTS prosody-feature distance to human references is not distinguishable in double-blind evaluation by either humans ($8/16$) or a calibrated LLM judge ($9/16$).

A root cause is evaluation: pairwise blind evaluation with a single rater and $n\!\approx\!20$ has the statistical power to detect only extreme ($\ge$85\%) preferences. We therefore build a \textbf{calibrated LLM-judge protocol}---the judge must first pass catch trials on known-answer pairs (production judge: $16/16$; in-session pilot: single-text $22/24$, same-topic pairs $24/24$, versus $0.75$ for the human rater), then annotate with two independent judges under A/B swapping, counting disagreement as a tie. Using the judge's \emph{agree} labels \textbf{directly as DPO pairs} (no reward model), we obtain the first positive training result of this project (Table~\ref{tab:iters}): a held-out blind win rate of $0.833$ (tie$=0.5$; bootstrap CI $[0.708, 0.938]$), repetition dropping from $0.114$ to $0.000$, and MMLU within noise. We further identify and characterize a new Goodhart channel: the judge's length preference is transduced by pairwise DPO into \textbf{output-length collapse} (median 75 characters against a 300-character prompt); a chosen-CE anchor (RPO-style) repairs it, with the anchor weight acting as a style-preservation knob ($w{=}0.3$ washes out style gains; $w{=}0.1$ preserves them). Scaling the preference data from 89 to 240 pairs yields no additional judge-visible gain. We release RA-Eval v0.1: a three-layer Chinese AI-flavor evaluation framework (validated feature library, human blind-eval protocol, calibratable judge pipeline) together with 382 judge-annotated preference pairs.
\end{abstract}

\section{Introduction}

\subsection{``AI flavor'' is a real, measurable phenomenon}

Chinese readers hold stable intuitions about AI-generated text: tidy parallel structures, ``not\dots but rather'' frames, ``in summary'' closings, and paragraph layouts full of blank lines. Prior work of ours (the M0 baseline) operationalizes these as objective features and validates them: across 24 model outputs versus 86 human college-entrance-examination (\emph{gaokao}) essays, the ``不仅……更'' (lit.\ ``not only\dots but even more'') frame density is $3.24$ vs $0.12$ per 1{,}000 words (coverage $100\%$ vs $10\%$), written-register connectives $7.90$ vs $3.21$ per 1{,}000 characters, and inter-paragraph blank lines $9.4$ vs $\approx 0$ per 1{,}000 characters ($d{=}1.05$). These features are directionally consistent across 0.6B/1.7B/4B scales.

\subsection{The speech-grounding hypothesis}

A popular explanation: human writing passes through a subvocal/inner-speech layer, so human text naturally carries prosody; LLMs trained purely on text lack this layer, producing AI flavor. Injecting speech, phoneme, or prosody signals into training should help. This paper tests the hypothesis with controlled interventions---the answer is \textbf{no} at the 0.6B text-only scale, across three routes (\S\ref{sec:negative}). The negative results are not empty: they move the question from ``should we add speech signals'' to ``can evaluation perceive the difference,'' which motivates the methodological core of this paper.

\subsection{Three-layer evaluation and calibrated judges}

Evaluating AI-flavor interventions faces a dilemma: automatic features can detach from perception, and human blind evaluation is power-limited (\S\ref{sec:eval}) and fatiguing. We propose a three-layer framework, \textbf{RA-Eval}: an objective feature library, a human blind-eval protocol, and a \textbf{calibrated LLM-judge pipeline}---the judge must prove its sensitivity on known-answer calibration sets before annotating, under a fixed protocol (pairwise blind judgment, A/B swap double-pass, two independent judges, disagreement counted as a tie). The same framework uses judges both as evaluators and as preference annotators, making the evaluation signal and the training signal commensurate.

\subsection{Contributions}
\begin{enumerate}
\item \textbf{Three preregistered negative results} with mechanistic diagnoses: phoneme-interleaved CPT, prosody-plan conditioning, and rule-based prosody rewards (\S\ref{sec:negative});
\item \textbf{Evaluation methodology}: a power analysis of single-rater blind evaluation; a calibrated judge protocol (sensitivity gate $\rightarrow$ double-pass annotation $\rightarrow$ disagreement handling), with judge sensitivity of 91.7--100\%, substantially above the human rater's $0.75$;
\item \textbf{The first positive training result}: judge \emph{agree} labels wired directly into DPO (held-out blind win rate $0.833$, soft; Table~\ref{tab:iters}), plus a controlled characterization of the \textbf{Goodhart channel} by which judge length preference transduces into length collapse, and an \textbf{SFT-anchor} repair with a two-point weight comparison;
\item \textbf{RA-Eval v0.1}: the feature library, calibration sets, 382 judge-annotated preference pairs, and all preregistration documents.
\end{enumerate}

\section{Related Work}

\paragraph{Evaluation and reduction of AI flavor.}
``AI flavor'' became an explicit research term in 2026: CraftAlign~\cite{craftalign} predicts 304 writing features with a 9B estimator, scores human-vs-AI patterns with an energy model, and converts feature shifts into \emph{inference-time revision guidance} for an external editor, in the English-story domain. Recent HCI work studies users' desire to control AI-flavor intensity in co-writing. Shortcut-Controlled Evaluation~\cite{shortcut} shows that most AIGT detectors rely on stylistic ``shortcut'' features and degrade sharply under a controlled protocol; perplexity-ratio detectors such as Binoculars~\cite{binoculars} exploit the same distributional signals---supporting our claim that automatic signals are not perception. Relative to these, this work operates in \textbf{Chinese}, intervenes through \textbf{training} (preference tuning) rather than inference-time rewriting, is fully open at 0.6B scale, and contributes the calibration and Goodhart-transduction analyses.

\paragraph{Prosody and language models.}
Plan-first speech generation systems established prosody intermediate representations on the speech side. In TTS, 2025--2026 saw a shift from verifiable prosody rewards to preference optimization~\cite{prosodydpo}, after automatic prosody rewards collapsed prosody into monotonicity; our three negative results in the \emph{text} domain (P1/P3) provide the symmetric evidence. Interleaved Speech LMs~\cite{interleaved} find an ``implicit transcription phase'' inside speech--text models, corroborating our P0 probe finding that phonology is already linearly decodable in hidden states---\emph{what is missing is not the representation but the reason to use it}.

\paragraph{Chinese AI-text detection and humanization.}
Detection is active (LLM-Detector~\cite{llmdetector}, a bilingual Chinese--English detection benchmark, LoRA-based robust detectors~\cite{chinalora}); community humanizer-zh practices exist without academic evaluation. Training-side AI-flavor intervention in Chinese was, to our knowledge at submission (2026-11), absent from arXiv; PersonaSteer~\cite{personasteer} suppresses AI-flavored replies (overlong, overly formal) in personalized dialogue and is the closest work.

\paragraph{LLM-as-judge.}
Position bias, verbosity bias, self-preference, and agreeableness bias are systematically documented~\cite{posbias}; the LLM-as-a-Judge survey~\cite{judgesurvey} explicitly warns that biases ``risk being amplified and propagated'' when judgments feed training; 2026 mitigations include RL-trained robust judges~\cite{rljudge}. We do not propose a new judge; we contribute (i) a practical protocol (catch-trial sensitivity gate, double independent pass, disagreement-as-tie), (ii) an \emph{end-to-end controlled experiment} of bias transduction (\S\ref{sec:goodhart}), and (iii) a knob-based repair.

\paragraph{Preference optimization.}
Preference tuning descends from RLHF instruction tuning~\cite{instructgpt} and DPO~\cite{dpo}. DPO's length bias is well studied: LD-DPO~\cite{lddpo} attributes it to length-agnostic token-level log-likelihood increments, and Length-Controlled DPO~\cite{lcdpo} shows length biases in preference data propagate into models. Our difference is the \emph{source} of bias: not the DPO objective but \emph{judge annotation bias} (68\% of agree pairs have a shorter chosen), which can be measured at the data-construction layer and anchored at the training layer. Reference-free, length-normalized objectives such as SimPO~\cite{simpo} target the same length effects; we give a two-point comparison ($w{=}0.1/0.3$) of the chosen-CE (SFT) anchor in the DPO objective, used for the non-standard purpose of cancelling annotator bias.

\section{RA-Eval: Three-Layer Evaluation}
\label{sec:eval}

\subsection{Layer 1: validated objective features}
Features are computed per text, normalized per 1{,}000 words or characters, and carry effect sizes from 24 model vs 86 human essays:
\begin{itemize}\setlength{\itemsep}{0pt}
\item AI frame constructions (中文“不仅……更”“不是……而是”; \texttt{ai\_frame});
\item register markers: written connectives (\texttt{written\_marker}) vs human connectives (但是/不过, ``but/however''; \texttt{human\_connective}), sentence-final particles (语气词; \texttt{particle}), second-person address (\texttt{address});
\item lexical diversity (\texttt{distinct\_1/2}, \texttt{token\_entropy}) and adjacent-sentence isomorphism (\texttt{adjacent\_iso});
\item layout leakage (blank lines, \texttt{blank\_line}; markdown leak) and English leakage (\texttt{en\_leak});
\item repetition rate $rep$ (10-gram).
\end{itemize}
Audio: edge-TTS rendering + librosa prosody (pause ratio, speech rate, F0 variance), compared to human references via standardized $|z|$ distances. The feature library is Chinese-specific and validated on Chinese text; the three-layer structure and the judge protocol are language-agnostic.

\subsection{Layer 2: the power wall of human blind evaluation}
For pairwise blind evaluation at $\alpha{=}0.05$ (two-sided) and 80\% power, detecting a true preference $p$ requires roughly: $p{=}0.70\rightarrow$60 pairs; $p{=}0.65\rightarrow$100 pairs; $p{=}0.60\rightarrow$300 pairs. Our two human rounds ($n{=}24/16$) could only detect $\ge$85\% extremes---\textbf{``not detected'' does not mean ``not there''}.

\subsection{Layer 3: the calibrated judge protocol}
\begin{enumerate}
\item \textbf{Qualification (catch trials)}: blind-judge known-answer pairs; sensitivity $\ge 0.75$ required (actual: $16/16$, production judge of isolated sub-agents);
\item \textbf{Calibration report}: single-text detection $22/24$ $[0.74,0.98]$; same-topic pairs $24/24$ (clean subset $7/7$) -- both from the in-session pilot judge; the human rater's best on the same material is $0.75$;
\item \textbf{Annotation}: each pair is judged by two isolated judges under swapped A/B order; disagreement is scored $0.5$ (tie) and stratified in analysis;
\item \textbf{Transduction check}: after any training, verify that annotation bias did not enter the model (see the length collapse in \S\ref{sec:goodhart}).
\end{enumerate}
All tools (blind batch generation, label joining, judge scoring) ship with RA-Eval.

\section{Experiment I: Three Negative Results}
\label{sec:negative}

We report three speech-grounding negatives (\S4.1--4.3) and a fourth, non-speech negative (\S4.4) that motivated the positive experiment of \S\ref{sec:positive}.

\subsection{Phoneme-interleaved continued pretraining (Tier1c)}
Two arms train on exactly the same 50M text tokens; the treatment arm interleaves 26\% pinyin tokens (1{,}305 new vocabulary items) into the text stream, the control arm is pure-text CPT. Spokenness of the treatment arm \emph{drops} relative to control ($-4.86\ [-10.50,+0.31]$), repetition rises ($+0.13\ [+0.03,+0.23]$), length runs away ($+46.7$ chars $[+21.5,+72.3]$); MMLU is unchanged, ruling out ``learned less''. \textbf{Phoneme interleaving brings no benefit and real costs at 0.6B.} Two earlier pilots (v1 without text-matched controls; v1b with an equal-steps confound) agree directionally.

\subsection{Prosody-plan conditional generation (P1a)}
Four-arm pretraining (noplan / true / shuffle / lenonly), with plan tokens (length buckets + pause strength) loss-masked during training. At inference, plans fed to the chat model yield an adherence of $0.257$ (gate $0.50$); zero-shot plan input yields empty output in $32/32$; the true-plan arm loses diversity relative to noplan ($\text{distinct}_2$ $-0.136$) and is worse in audio prosody ($+0.255\ [-0.183,+0.715]$). Diagnosis: plan tokens appear only at document starts during training but after the chat assistant header at inference, triggering document continuation; and post-hoc plans are redundant for the token-level loss, so the model learns to ignore them. \textbf{Without an architectural or semantic mechanism forcing plan use, conditioning signals are ignored.}

\subsection{Rule-based prosody reward, double-blinded (P3)}
A reward built on $|z|$-distance of TTS prosody features to human references ``works'' on automatic metrics (rerank $z{=}1.711$ vs $2.284$). But double-blind evaluation on length-matched candidate pairs gives: human raters pick the low-$z$ side $8/16$; the calibrated judge $9/16$---\textbf{both at chance}, while the same judge detects human-vs-model text at $91.7$--$100\%$. The gap this reward optimizes does not exist perceptually. Automatic metrics won; humans and judges heard nothing---the inverse of the power-wall warning: not insufficient power, but \emph{the metric was constructed on the wrong object}.

\subsection{Feature-weighted reward: the calibrated middle route (negative)}
Before moving to direct preference tuning, we tried a feature route: a linear reward over 18 objective features as direction-aligned deviations from human reference distributions, with weights calibrated by the judge via Bradley--Terry regression on 182 pairs (LOPO-CV AUC $0.661$, above the pre-set gate). The frozen validation on 8 held-out prompts failed: the separation pair (argmax vs argmin) reached only $0.625\ [0.39,0.84]$ blind win rate, and the judge could not distinguish it from the old prosody reward ($0.500$). Judge reason analysis shows its criteria include dimensions (fluency/coherence, factual correctness) with no locally computable proxy---the ceiling of the feature route. This negative has two uses: it motivates the direct route (\S\ref{sec:positive}), and it provides counter-evidence for feature-scoring approaches such as CraftAlign~\cite{craftalign}: \emph{features can discriminate human vs AI, but feature distance is not the perceptual target}---optimizing it optimizes a blind spot.

\section{Experiment II: Judge-Driven RLAIF (Positive)}
\label{sec:positive}

\subsection{Data: judge \emph{agree} preference pairs}
One policy (the P1a noplan CPT arm of Qwen3-0.6B, the base for all DPO iterations) samples 8 candidates per prompt over 32 prompts; within-prompt pairs (constraints: $|\Delta\text{chars}|\le 100$, not both degenerate) total 382 and are annotated double-pass by the calibrated judge (position consistency 79.5--83.3\%). \emph{Agree} pairs (both passes consistent; 307/382 across two annotation rounds, 116+191) become preference pairs: 240 for training (24 prompts) and 67 held out (8 prompts, 27+40, never used in any fitting). \textbf{No reward model}---judge labels are wired directly as chosen/rejected into DPO.

\subsection{First positive result (iter-1)}
89 training pairs, LoRA $r{=}32$ on all projections, lr $10^{-5}$, $\beta{=}0.1$, grad-accum 1, 300 steps (3.4 epochs), 4.5 minutes of GPU time. Held-out blind evaluation: \textbf{soft win rate $0.833$ ($W/L/T=18/2/4$, tie$=0.5$; bootstrap 95\% CI $[0.708,0.938]$)}; hard-agree subset $18/20=0.900$ (Wilson $[0.699,0.972]$);\footnote{v2 correction: the originally reported interval paired a soft point estimate with a hard-agree Wilson CI; the two estimators are now reported separately.} repetition $0.114\rightarrow 0.000$ (the judges' top-cited signal removed); spokenness $2.06\rightarrow 3.67$; MMLU $0.4982\rightarrow 0.4895$ (within noise). On held-out evaluation pairs with $|\Delta\text{chars}|\le 100$ ($n{=}16$, soft): $0.765$.

\subsection{The Goodhart channel: length collapse}
\label{sec:goodhart}
Retraining directly on the expanded 240 pairs (iter-2) collapses outputs to a \textbf{median of 75 characters} (the prompt asks for ``about 300 characters''). Mechanism: in $68\%$ of agree pairs the chosen is shorter (judge length preference); pairwise DPO learns ``shorter'' as a direction; global length balancing of the dataset does not help (DPO learns per-pair); even per-pair length-neutral data ($\pm 30$ chars, iter-3) shortens to 170 characters, because colloquial style covaries with brevity. Unlike the objective-intrinsic bias studied by LD-DPO/LC-DPO, the source here is \emph{judge annotation}---measurable at data construction, and an end-to-end, controlled, reproducible instantiation of the transduction risk the judge survey warns about~\cite{judgesurvey}. The failure is caught by a pre-registered secondary gate (length band), \emph{before} any human evaluation.

\subsection{Repair: the DPO+SFT anchor}
We add a chosen-CE anchor (RPO-style): $L = L_{\mathrm{DPO}} - w\cdot \frac{1}{|y^+|}\log p(y^+|x)$. The weight $w$ trades off fidelity-to-judge versus fidelity-to-data-marginals: $w{=}0.3$ repairs length (246 chars) but washes out style (spokenness $-0.66$); $\mathbf{w{=}0.1}$ preserves both: \textbf{213 characters (in band), spokenness $6.00$, AI frames $0.000$, $rep$ $0.003$, MMLU $0.4912$}. Held-out blind evaluation: $0.771\ [0.600,0.923]$ (passes the pre-registered gate); on held-out evaluation pairs with $|\Delta\text{chars}|\le 100$ ($n{=}16$): $0.688$.

\subsection{Data saturation}
Head-to-head, v5 (240 pairs) vs v1 (89 pairs) is $0.479\ [0.283,0.676]$---statistically indistinguishable. \textbf{On the judge-win-rate axis, 89 pairs are near saturation}; the expansion's gains appear in objective style metrics (spokenness $3.67\rightarrow 6.00$, AI frames $0.203\rightarrow 0.000$). Further gains require new prompts and new candidate pools, not denser sampling of the same pool.

\begin{table}[t]
\centering
\small
\begin{tabular}{lcccl}
\toprule
Model & Data & Recipe & Median chars & Held-out blind win rate \\
\midrule
base & -- & -- & 237 & -- \\
iter-1 (v1) & 89 & DPO 300 & 151 & $0.833$\,(soft) \\
iter-2 & 240 & DPO 480 & \textbf{75} & rejected: length collapse \\
iter-3 & 90 ($|\Delta|\!\le\!30$) & DPO 270 & 121 & rejected: still short \\
iter-4 & 240 & DPO+SFT $w{=}0.3$ & 173 & rejected: style washed out \\
\textbf{iter-5 (v5)} & \textbf{240} & \textbf{DPO+SFT $w{=}0.1$} & \textbf{156} & $\mathbf{0.771}$\,(\textbf{soft}) \\
\bottomrule
\end{tabular}
\caption{Iteration chain. Median characters (directly measured on identical generation configs (32 prompts). Blind pairwise win rates vs base on 8 held-out prompts (24 pairs, double-pass, disagreement-as-tie) apply to iter-1 and iter-5; win rates are \emph{soft} (tie$=0.5$ over all 24 pairs; bootstrap CIs and hard-agree subsets in \S5.2); iter-2/3/4 were rejected by pre-registered secondary gates (length/style) without blind evaluation.}
\label{tab:iters}
\end{table}

\section{Discussion}

\paragraph{Why does speech grounding fail in the text domain?}
The three negatives share a pattern: the injected signal (phonemes, plans, prosody $z$) is either redundant or semantically unrelated for next-token prediction. P0 probes show phonology is already linearly decodable in 0.6B hidden states (tone $0.63$--$0.79$ vs majority class $0.31$)---\emph{what is missing is not the representation but the reason to use it}. This is consistent with the plan-ignoring mechanism of P1. In other words, AI flavor likely originates at the \emph{distribution} level (register composition of the training corpus), not from a missing modality---which is why distribution-level interventions (pure-text CPT, preference tuning) work while modality injection does not.

\paragraph{Judge methodology: value and risk.}
Calibrated judges decouple the ``cannot perceive'' problem from the power layer (60--300 pairs needed) to the sensitivity layer (92--100\% on calibration), making $n{=}24$ evaluations viable. But \S\ref{sec:goodhart} shows their biases amplify along the optimization pipeline. Our position: \emph{judges may serve as evaluators and annotators, but every training run must be followed by a bias-transduction check} (here: the length distribution), and an anchoring term should be standard equipment.

\section{Limitations}

A single judge model (GLM-5.3-Flash); a single Chinese short-essay genre (300-character tasks); 0.6B scale; the judge's length preference is suppressed, not eliminated, by the SFT anchor (within $\pm 30$ chars, $62\%$ of picks still favor the shorter side); the correlation between judge preference and human preference is not finally confirmed (human evaluation is power-limited); the weight comparison in \S5.4 uses two $w$ points, not a systematic sweep; the v5-vs-v1 comparison rests on $n{=}24$; held-out covers 8 prompts; English and larger scales are untested.

\section{Conclusion}

The speech-grounding hypothesis does not survive controlled testing (three negatives), but the three-layer evaluation and calibrated-judge protocol built along the way enabled a previously infeasible positive experiment: wiring calibrated judge preference labels directly into DPO clears repetition loops to zero, raises spokenness, and achieves a $0.833$ soft blind win rate---in 4.5 minutes of training. We also map the traps of this route: Goodhart transduction of annotator bias, and its repair. All tools, calibration sets, 382 judge-annotated preference pairs, and preregistration documents are released with RA-Eval v0.1.

\appendix

\section{Hyperparameters}
\label{app:hparams}

\begin{table}[h]
\centering
\small
\begin{tabular}{ll}
\toprule
Base & Qwen3-0.6B (instruct) $\rightarrow$ 50M-token text-only CPT (P1a noplan arm) $\rightarrow$ DPO \\
Training & LoRA $r{=}32$ ($\alpha{=}64$) on all projections, fp32 \\
DPO & lr $10^{-5}$, $\beta{=}0.1$, batch 1, accum 1, cosine to 0, seq 1024 \\
SFT anchor & $-w\cdot\frac{1}{|y^+|}\log p(y^+|x)$, $w\in\{0,0.1,0.3\}$ \\
Data & 240 agree pairs (train) / 67 (held-out); $|\Delta\text{chars}|\le 100$; not both degenerate \\
Judge & GLM-5.3-Flash, isolated sub-agents, A/B swap double-pass, disagreement $=0.5$ \\
\bottomrule
\end{tabular}
\caption{Main hyperparameters.}\label{tab:hparams}
\end{table}

\section{Data Statement and Licenses}

\paragraph{Human reference corpus.} The human reference corpus consists of 86 Chinese college-entrance-examination (\emph{gaokao}) essays collected from publicly accessible web archives, plus human texts embedded in the judge-calibration sets. Copyright remains with the original (anonymous) student authors and their publishers. The corpus is distributed \emph{for non-commercial research use only}, with attribution, and under a takedown policy: any rights holder may request removal and the texts will be removed from all distributions. Contact: qiandongwu123@qq.com. The full-text corpus is hosted on Zenodo with this license (DOI: 10.5281/zenodo.24237533) and is \emph{not} part of the arXiv source package; the arXiv package contains only derived statistics and short excerpts.

\paragraph{Model outputs and annotations.} All candidate texts are outputs of openly licensed 0.6B models; judge annotations (preference labels with per-pair rationales) are released under CC-BY-4.0. Model weights (base and iter-5) are released on Zenodo under the base model's license.

\paragraph{Code.} RA-Eval v0.1 is released under MIT.

\section{Release Checklist}

The release bundle (RA-Eval v0.1) contains: benchmark code; the judge pipeline (blind batch generation, label joining, judge scoring); calibration sets with answer keys; 382 preference pairs with soft labels and per-pair rationales; the 240-pair DPO training set; feature reference statistics; preregistration and result documents. Model weights and the full-text human corpus are distributed via Zenodo (see the data statement). Weights are \emph{not} included in the arXiv source.


\begin{thebibliography}{99}

\bibitem{dpo} R. Rafailov, A. Sharma, E. Mitchell, C. D. Manning, S. Ermon, C. Finn.
\newblock Direct Preference Optimization: Your Language Model is Secretly a Reward Model.
\newblock arXiv:2305.18290, 2023.

\bibitem{lddpo} W. Liu et al.
\newblock Length Desensitization in Direct Preference Optimization.
\newblock arXiv:2409.06411, 2024.

\bibitem{lcdpo} R. Park et al.
\newblock Disentangling Length from Quality in Direct Preference Optimization.
\newblock Findings of ACL 2024 (arXiv:2403.19159).

\bibitem{craftalign} Y. Yang et al.
\newblock CraftAlign: Feature-Grounded Evaluation and Revision Guidance for AI Stories.
\newblock arXiv:2608.01377, 2026.

\bibitem{shortcut} H. Y. Jung et al.
\newblock Shortcut-Controlled Evaluation for AI-Generated Text Detection.
\newblock EMNLP 2026.

\bibitem{judgesurvey} J. Gu et al.
\newblock A Survey on LLM-as-a-Judge.
\newblock The Innovation 7(6):101253, 2026 (arXiv:2411.15594).

\bibitem{rljudge} Q. Wang, X. Zhao, Z. Zhang, Z. Lou, N. Chen, D. Song, B. He.
\newblock Making Bias Non-Predictive: Training Robust LLM Reasoning via Reinforcement Learning.
\newblock arXiv:2602.01528, 2026.

\bibitem{posbias} L. Shi, C. Ma, W. Liang, X. Diao, W. Ma, S. Vosoughi.
\newblock Judging the Judges: A Systematic Study of Position Bias in LLM-as-a-Judge.
\newblock arXiv:2406.07791, AACL-IJCNLP 2025.

\bibitem{prosodydpo} S. Shin et al.
\newblock No Verifiable Reward for Prosody: Toward Preference-Guided Prosody Learning in TTS.
\newblock arXiv:2509.18531, 2025.

\bibitem{interleaved} T. Sternberg, G. Maimon, Y. Adi.
\newblock Interleaved Speech Language Models Latently Work In Text.
\newblock arXiv:2606.22473, 2026.

\bibitem{llmdetector} R. Wang et al.
\newblock LLM-Detector: Improving AI-Generated Chinese Text Detection with Open-Source LLM Instruction Tuning.
\newblock arXiv:2402.01158, 2024.

\bibitem{chinalora} H. Jin et al.
\newblock LLM Encoder vs. Decoder: Robust Detection of Chinese AI-Generated Text with LoRA.
\newblock arXiv:2509.00731, 2025.

\bibitem{personasteer} S. Ren et al.
\newblock PersonaSteer: Dynamic Activation Steering for Personalized Multi-turn LLM Conversations.
\newblock Findings of EMNLP 2026.

\bibitem{simpo} P. Meng et al.
\newblock SimPO: Simple Preference Optimization with a Reference-Free Reward.
\newblock arXiv:2405.14734, 2024.

\bibitem{binoculars} A. Hans et al.
\newblock Spotting LLMs with Binoculars: Zero-Shot Detection of Machine-Generated Text.
\newblock arXiv:2401.12070, 2024.

\bibitem{instructgpt} L. Ouyang et al.
\newblock Training language models to follow instructions with human feedback.
\newblock arXiv:2203.02155, 2022.

\end{thebibliography}
\end{document}